\documentclass{article}

\usepackage[preprint]{neurips_2026}

\usepackage[utf8]{inputenc}
\usepackage[T1]{fontenc}
\usepackage{hyperref}
\usepackage{url}
\usepackage{booktabs}
\usepackage{amsfonts}
\usepackage{amsmath}
\usepackage{nicefrac}
\usepackage{microtype}
\usepackage{xcolor}
\usepackage{graphicx}

\newcommand{\tdaacc}{0.904}

\title{HyperShadow: A Benchmark for Detecting 3D Projections of Higher-Dimensional Spatial Objects}

\author{%
  Akshay Sasi \\
  Independent Researcher \\
  \texttt{akshaysasi12.knr@gmail.com} \\
}

\begin{document}

\maketitle

\begin{abstract}
Machine-learning datasets labelled ``4D'' universally denote three spatial dimensions plus time. We introduce \textbf{HyperShadow}, the first public benchmark in which the fourth, fifth, and sixth dimensions are \emph{spatial}: the task is to decide whether a 3D point cloud is a native three-dimensional shape or the projection, the ``shadow'', of a rigid object living in $\mathbb{R}^N$ ($N{=}4$--$6$). We show this task is fundamentally distinct from intrinsic-dimension estimation: a shadow is still at-most-3-dimensional data, and standard estimators (TwoNN, Levina--Bickel MLE) reach only 71--73\% accuracy. Detection instead requires \emph{projection signatures}, density folds, filled volumes with characteristic radial profiles, and topology changes, which a 190k-parameter point network recovers at 96.2\% accuracy (std 0.3\% over five seeds) across four corruption tiers, generalizing at 79--91\% to object families never seen in training. On a temporal track of rigidly rotating objects we introduce a zero-parameter \emph{rigidity witness}: the residual of the optimal rigid 3D alignment (Kabsch) between consecutive frames, which must vanish for any rigid 3D motion but cannot vanish for the shadow of a rigid rotation in $\mathbb{R}^N$. This single interpretable statistic separates the classes at AUROC 0.982. All data are generated reproducibly from seeds; the dataset, models, and code are released publicly. HyperShadow makes no claim about physical reality; it is a controlled instrument for studying which observable statistics can \emph{certify incompatibility with a purely three-dimensional explanation}.
\end{abstract}

\section{Introduction}

A cube held before a lamp casts a two-dimensional shadow. The shadow is genuinely 2D, every point of it lies in the plane, yet it is not a \emph{generic} 2D shape: as the cube rotates, the shadow deforms in ways no rigid flat object could, and even a single frame carries statistical traces of the projection that produced it. This paper asks the analogous question one dimension up, computationally: \textbf{given only a 3D point cloud, or a short sequence of them, can any method decide whether the data is the projection of a higher-dimensional rigid object?}\footnote{Code and tests: \url{https://github.com/AkshaySasi/hypershadow}. Dataset: \url{https://huggingface.co/datasets/AkshaySasi/hypershadow}. Trained models: \url{https://huggingface.co/AkshaySasi/hypershadow-models}.}

Three observations motivate the benchmark.

\paragraph{(1) The task is not intrinsic-dimension estimation.} A rich literature estimates the dimension of the manifold underlying observed data \citep{levina2005mle,facco2017twonn,lee2025survey}. But a projection \emph{reduces} dimension: the orthographic shadow of the 3-sphere $S^3 \subset \mathbb{R}^4$ is a solid 3-ball; the shadow of the 2-dimensional Clifford torus remains 2-dimensional. Intrinsic dimension of the observed data therefore cannot, even in principle, fully identify shadows. What distinguishes them is \emph{how} mass, density, and topology are arranged, the signatures of the projection map itself. Section~\ref{sec:results-static} quantifies this gap: classical estimators reach 73\% on HyperShadow where a small learned model reaches 96\%.

\paragraph{(2) ``4D'' datasets in machine learning are space + time.} Existing 4D benchmarks, 4D world models and dynamic point-cloud suites such as MSR-Action3D, DeformingThings4D, and HOI4D \citep{tesseract2025,omniworld2025}, model 3D geometry evolving over time. To our knowledge no public dataset contains projections of objects with four or more \emph{spatial} dimensions, despite the pedagogical ubiquity of the tesseract.

\paragraph{(3) Humans can do a version of this task.} \citet{he2023rigidity} showed in psychophysical experiments that human observers discriminate rigid from non-rigid motion of a hypercube's projection at accuracy comparable to the 3D case, suggesting the perceptual system exploits geometric regularities that do not require four-dimensional experience. There is, however, no machine counterpart: no dataset, no baseline, no algorithmic detector. HyperShadow provides all three, and our rigidity witness (Section~\ref{sec:witness}) is the explicit algorithmic analogue of the human judgement studied there.

\paragraph{Contributions.}
\begin{enumerate}
  \item \textbf{HyperShadow}, a reproducible benchmark of 10{,}800 static point clouds and 1{,}800 temporal sequences: native 3D shapes vs.\ 3D projections of eleven object families in $\mathbb{R}^4$--$\mathbb{R}^6$, under two projection models and four cumulative corruption tiers, with fairness rules that remove every shortcut we could identify (Section~\ref{sec:benchmark}).
  \item \textbf{A characterization of the failure of intrinsic-dimension estimation} on projected data, establishing that shadow detection is a distinct problem (Section~\ref{sec:results-static}).
  \item \textbf{Baselines spanning three methodological families}, hand-crafted geometric features, persistent homology, and a compact learned point network, including leave-one-family-out generalization tests (Sections~\ref{sec:baselines}, \ref{sec:results}).
  \item \textbf{The rigidity witness}: a zero-parameter, closed-form statistic for temporal data that certifies incompatibility with rigid 3D motion, achieving AUROC 0.982 (Section~\ref{sec:witness}). We propose \emph{dimensional witnesses}, statistics whose value is provably bounded under any 3D explanation, as a general template, analogous in logical structure to Bell inequalities: one does not observe the hidden structure, one rules out the class of explanations that lack it.
\end{enumerate}

\paragraph{Scope.} All results hold under the stated simulation assumptions. Nothing in this paper is evidence for or against higher spatial dimensions in physical reality; the benchmark is an instrument for studying detectability, validated where ground truth is known.

\section{Related work}

\paragraph{Intrinsic-dimension estimation.} Estimators from Levina--Bickel MLE \citep{levina2005mle} to TwoNN \citep{facco2017twonn} and their successors, surveyed by \citet{lee2025survey}, infer the dimension of the data manifold, typically benchmarked on manifolds embedded in \emph{higher}-dimensional ambient spaces \citep{bac2021scikit}. HyperShadow inverts the setting: the generating object has higher dimension than the observation space. Learned bottleneck approaches such as IDEA \citep{idea2025} estimate dimension with autoencoders; they too target the data manifold, not the generator.

\paragraph{Topological data analysis.} Persistent homology summarizes multi-scale topology of point clouds \citep{tralie2018ripser}, with recent work making $H_1$/$H_2$ computation robust in high ambient dimension \citep{spectral2024} and enriching standard 3D benchmarks with topological features \citep{tophshape2026}. Projection alters topology (shadows self-intersect and fill voids), which motivates our TDA baseline.

\paragraph{Point-cloud learning.} PointNet-style set networks \citep{qi2017pointnet} and successors dominate 3D point classification on ModelNet40 and ScanObjectNN \citep{uy2019scanobjectnn}. Our PointNet-lite is a deliberately small member of this family, sized to demonstrate that the task does not require large models or compute.

\paragraph{Four spatial dimensions.} Computer-graphics work renders 4D scenes for visualization \citep{hdgraphics2021}; vision science shows humans judge 4D rigidity from 3D projections \citep{he2023rigidity}. Existing ``4D'' ML datasets are dynamic 3D \citep{tesseract2025,omniworld2025}. None provide a benchmark of higher-spatial-dimension projections.

\section{The HyperShadow benchmark}
\label{sec:benchmark}

\begin{figure}
  \centering
  \includegraphics[width=\linewidth]{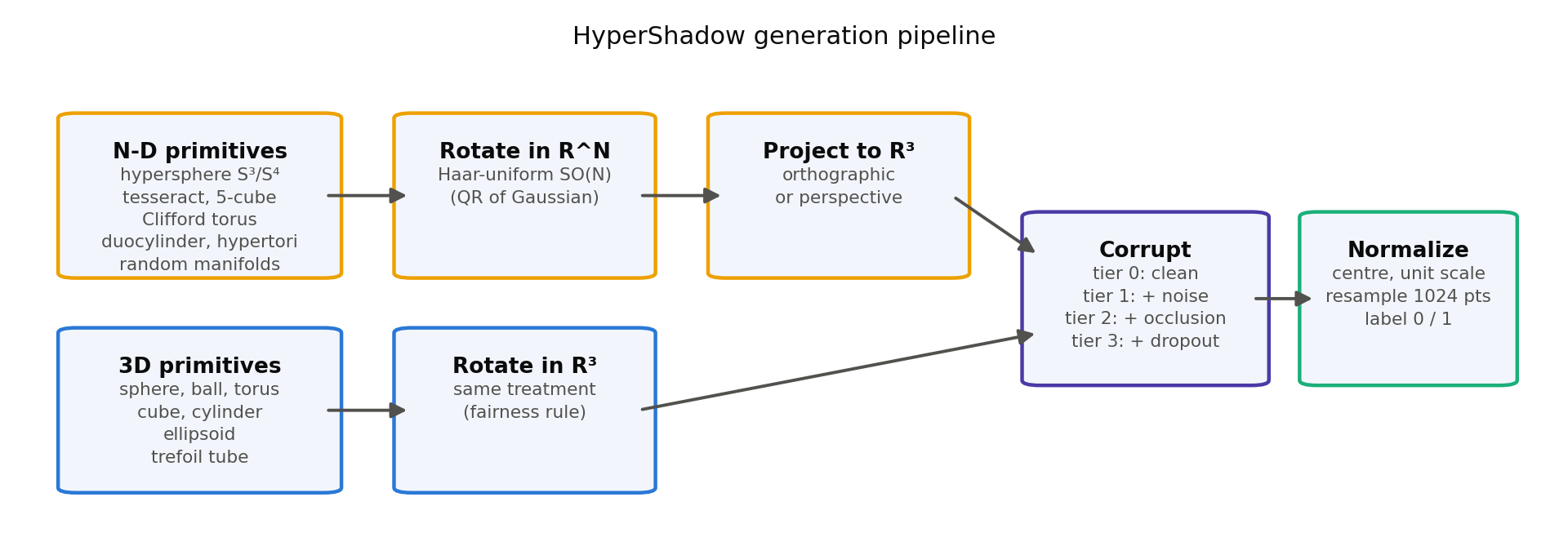}
  \caption{The HyperShadow generation pipeline. Higher-dimensional primitives are rotated Haar-uniformly in $\mathbb{R}^N$ and projected to $\mathbb{R}^3$; native 3D primitives receive the identical rotation and corruption treatment (fairness rule), so no preprocessing artifact separates the classes.}
  \label{fig:pipeline}
\end{figure}

Figure~\ref{fig:pipeline} summarizes generation. All code is seeded NumPy; every sample is reproducible from a seed and a metadata record.

\subsection{Object families}

\textbf{Native 3D (label 0), 7 families:} sphere, solid ball, torus (area-uniform via rejection sampling), cube surface, capped cylinder, random ellipsoids, and a tube around a trefoil knot (topologically non-trivial). Sampling is uniform with respect to surface area or volume.

\textbf{Higher-dimensional (label 1), 11 families:} unit hyperspheres $S^3$ and $S^4$; the solid 4-ball; tesseract and 5-cube boundaries (uniform over cells); the Clifford torus and duocylinder ridge in $\mathbb{R}^4$ (flat tori, uniform in angles); generalized hypertori in $\mathbb{R}^4$ and $\mathbb{R}^5$ (rejection-corrected); and random smooth compact manifolds embedded in $\mathbb{R}^5$/$\mathbb{R}^6$ by random trigonometric maps, providing unbounded family diversity.

\subsection{Projection and corruption}

Each $N$-D object receives a Haar-uniform rotation in $\mathrm{SO}(N)$ (QR decomposition of a Gaussian matrix with sign correction), then is projected to $\mathbb{R}^3$ \emph{orthographically} (drop coordinates, the mathematical shadow) or \emph{perspectively} (successive division through a virtual viewpoint along each extra axis, the 4D pinhole camera). Four cumulative corruption tiers follow: \textbf{tier 0} clean; \textbf{tier 1} Gaussian jitter ($\sigma = 2\%$ of cloud scale); \textbf{tier 2} $+$ removal of a random 20--40\% half-space slab (self-occlusion); \textbf{tier 3} $+$ heavier jitter and distance-dependent dropout from a random viewpoint (LiDAR-like sparsity).

\subsection{Fairness rules}
\label{sec:fairness}

Benchmarks of synthetic classes are vulnerable to shortcuts. We remove every one we identified:
\begin{itemize}
  \item \textbf{Identical treatment}: native shapes receive the same Haar-random (3D) rotations and identical corruption tiers.
  \item \textbf{Normalization}: every cloud is centred, scaled to unit mean radius, and resampled to exactly 1{,}024 points, no information in position, scale, or cardinality.
  \item \textbf{Volume is not the answer}: the solid ball sits in the native class, so ``fills a volume'' cannot solve the task; conversely the Clifford torus shadow remains a surface, so ``is a surface'' fails too.
  \item \textbf{Class balance by tier}: corruption tiers are distributed identically across classes.
\end{itemize}

\subsection{Tracks}

\textbf{Static}: 10{,}800 clouds of 1{,}024 points (600 per family), split 70/10/20 into train/validation/test. \textbf{Temporal}: 1{,}800 sequences of 16 frames $\times$ 512 points. A single object rotates rigidly at constant angular velocity along a random one-parameter subgroup of $\mathrm{SO}(d)$ (matrix exponential of a random antisymmetric generator); each frame is projected and normalized independently, with point identity preserved across frames. For label 0 the rigid rotation is in $\mathbb{R}^3$; for label 1, in $\mathbb{R}^N$.

\section{Baselines}
\label{sec:baselines}

\paragraph{Intrinsic dimension (0 parameters).} TwoNN \citep{facco2017twonn} and Levina--Bickel MLE \citep{levina2005mle} computed per cloud; a single decision threshold (direction and value fit on half the data, evaluated on the other half).

\paragraph{Geometric features + GBT.} 39 rotation-invariant features per cloud, the D2 pairwise-distance histogram, radial mass profile, PCA eigenvalue spectrum, $k$-NN distance-ratio statistics (a local dimension proxy), and nearest-neighbour density statistics, classified by gradient-boosted trees under 5-fold cross-validation.

\paragraph{Persistent homology + GBT.} Ripser \citep{tralie2018ripser} persistence diagrams up to $H_2$ on 256-point subsamples; per-dimension lifetime statistics and top-5 $H_1$/$H_2$ lifetimes; same classifier protocol.

\begin{figure}
  \centering
  \includegraphics[width=\linewidth]{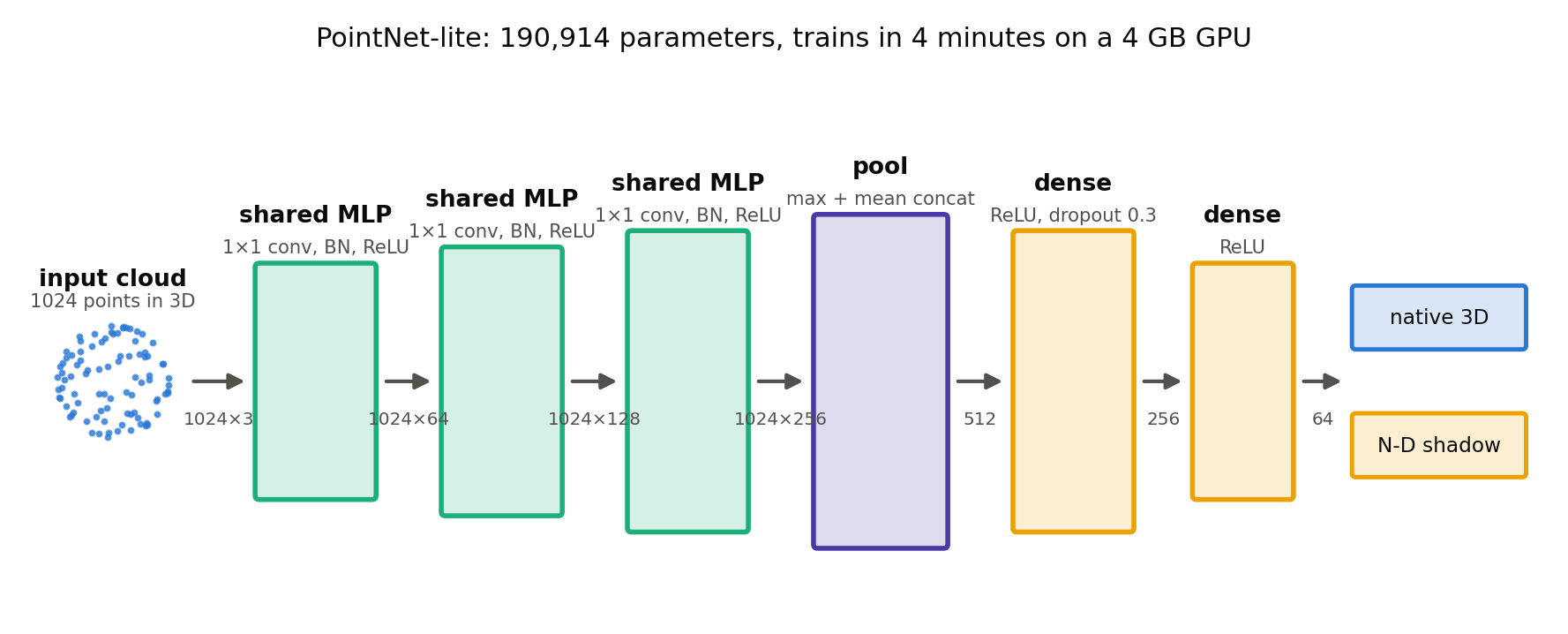}
  \caption{PointNet-lite. A compact permutation-invariant network suffices: shared per-point MLPs, symmetric pooling, and a two-layer head, 190k parameters, trained in minutes on a 4\,GB consumer GPU.}
  \label{fig:arch}
\end{figure}

\paragraph{PointNet-lite.} A 190{,}914-parameter set network (Figure~\ref{fig:arch}): shared per-point MLP (64--128--256 channels, batch normalization), concatenated max- and mean-pooling, and a small dropout-regularized head. Training: AdamW, cosine schedule, 60 epochs, random-rotation and jitter augmentation; approximately 4 minutes on a GTX 1650 Ti (4\,GB). The deliberate smallness is a claim: the signal is strong enough that no scale is needed.

\section{The rigidity witness}
\label{sec:witness}

Let $X_t \in \mathbb{R}^{n\times 3}$ be corresponding points across frames. For any rigid 3D motion there exist $R \in \mathrm{SO}(3)$, translation $t$, and scale $s$ (absorbing per-frame normalization) with $X_{t+1} = s X_t R^\top + t$ exactly; the optimal $(s, R, t)$ is closed-form (Kabsch/Procrustes). Define the witness
\begin{equation}
  w(X) \;=\; \frac{1}{T-1}\sum_{t=1}^{T-1} \operatorname{RMS}\bigl(X_{t+1} - \hat{s}_t X_t \hat{R}_t^\top - \hat{t}_t\bigr),
\end{equation}
the mean residual of the optimal rigid alignment between consecutive frames.

For native sequences $w$ is bounded by the noise floor. For the shadow of a rigid rotation in $\mathbb{R}^N$, the apparent 3D motion composes visible rotation with a re-weighting of hidden coordinates; no rigid 3D map explains it, and $w$ stays bounded away from zero. Crucially, $w$ is a \emph{certificate}: a large value does not merely correlate with higher dimensionality, it \emph{rules out} the entire class of rigid-3D explanations, up to noise. This is the logical structure of a Bell inequality, transplanted to kinematics.

\begin{figure}
  \centering
  \includegraphics[width=0.78\linewidth]{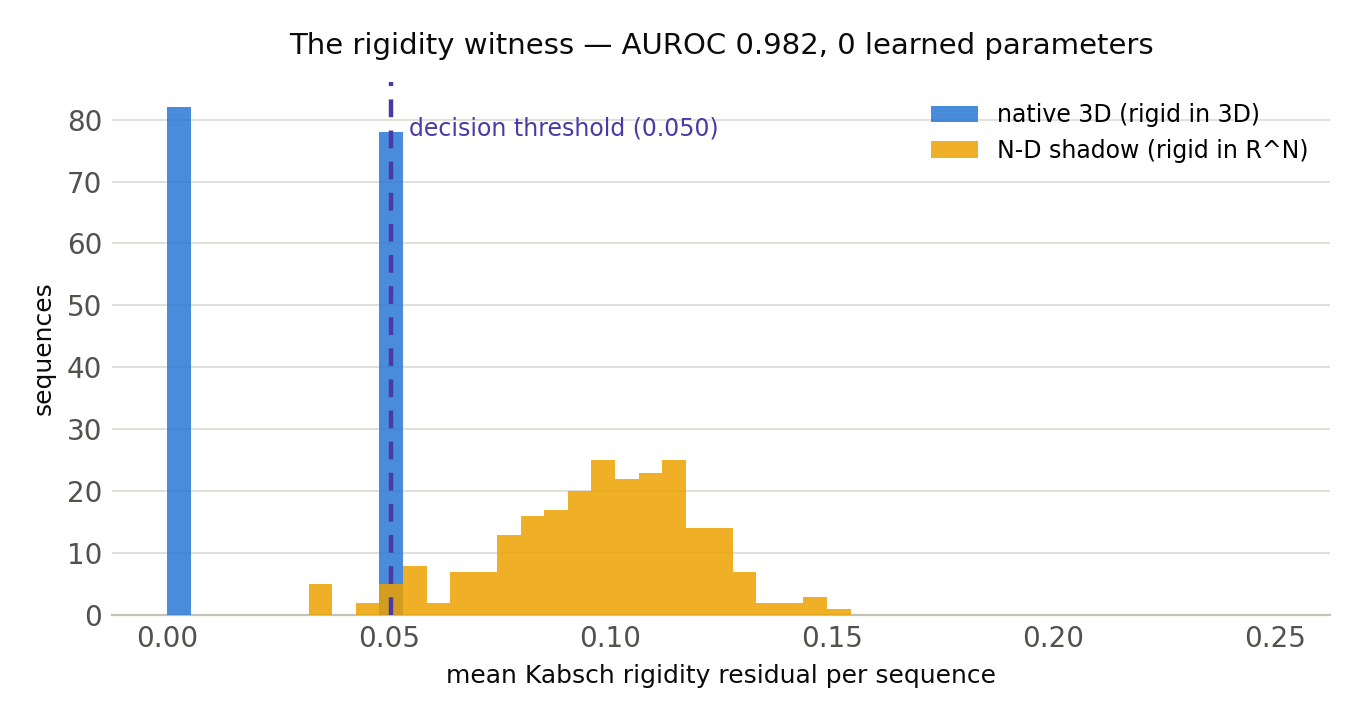}
  \caption{Distributions of the rigidity witness $w$. Native 3D motion (blue) is exactly rigid (clean tier, $w\approx 0$) or rigid up to the noise floor; shadows of rigid $\mathbb{R}^N$ rotations (yellow) cannot be rigidly explained in 3D. One threshold separates them at AUROC 0.982 with no learned parameters.}
  \label{fig:rigidity}
\end{figure}

Empirically (Figure~\ref{fig:rigidity}): native sequences concentrate at $w \approx 0$ (clean tier) and $w \approx 0.05$ (the noise tier's floor); shadows concentrate at $w \approx 0.10$, four times the clean-motion residual. A single threshold fit on half the sequences achieves \textbf{held-out accuracy $0.978 \pm 0.004$ over five random splits and AUROC 0.982}, with zero learned parameters. The hardest families (Clifford torus 0.92, tesseract 0.93) are those whose random rotation plane most often lies almost entirely within the visible subspace, making the shadow \emph{nearly} rigid, a geometric, not statistical, limitation.

\section{Results}
\label{sec:results}

\subsection{Static track}
\label{sec:results-static}

\begin{table}
  \caption{Static track: held-out accuracy for detecting whether a single 3D point cloud is the projection of a higher-dimensional object (chance $= 0.5$). Threshold methods and PointNet-lite report mean $\pm$ standard deviation over five random splits/seeds; the feature and homology baselines are 5-fold cross-validation. Intrinsic-dimension estimation is structurally unable to solve the task; projection signatures are learnable.}
  \label{tab:static}
  \centering
  \begin{tabular}{lcc}
    \toprule
    Method & Learned params & Accuracy \\
    \midrule
    Chance & --- & 0.500 \\
    TwoNN threshold \citep{facco2017twonn} & 0 & 0.718 $\pm$ 0.006 \\
    Levina--Bickel MLE threshold \citep{levina2005mle} & 0 & 0.737 $\pm$ 0.004 \\
    Persistent homology + GBT & --- & \tdaacc \\
    Geometric features + GBT & --- & 0.956 \\
    \textbf{PointNet-lite (ours)} & 190k & \textbf{0.962 $\pm$ 0.003} \\
    \bottomrule
  \end{tabular}
\end{table}

\begin{figure}
  \centering
  \includegraphics[width=0.9\linewidth]{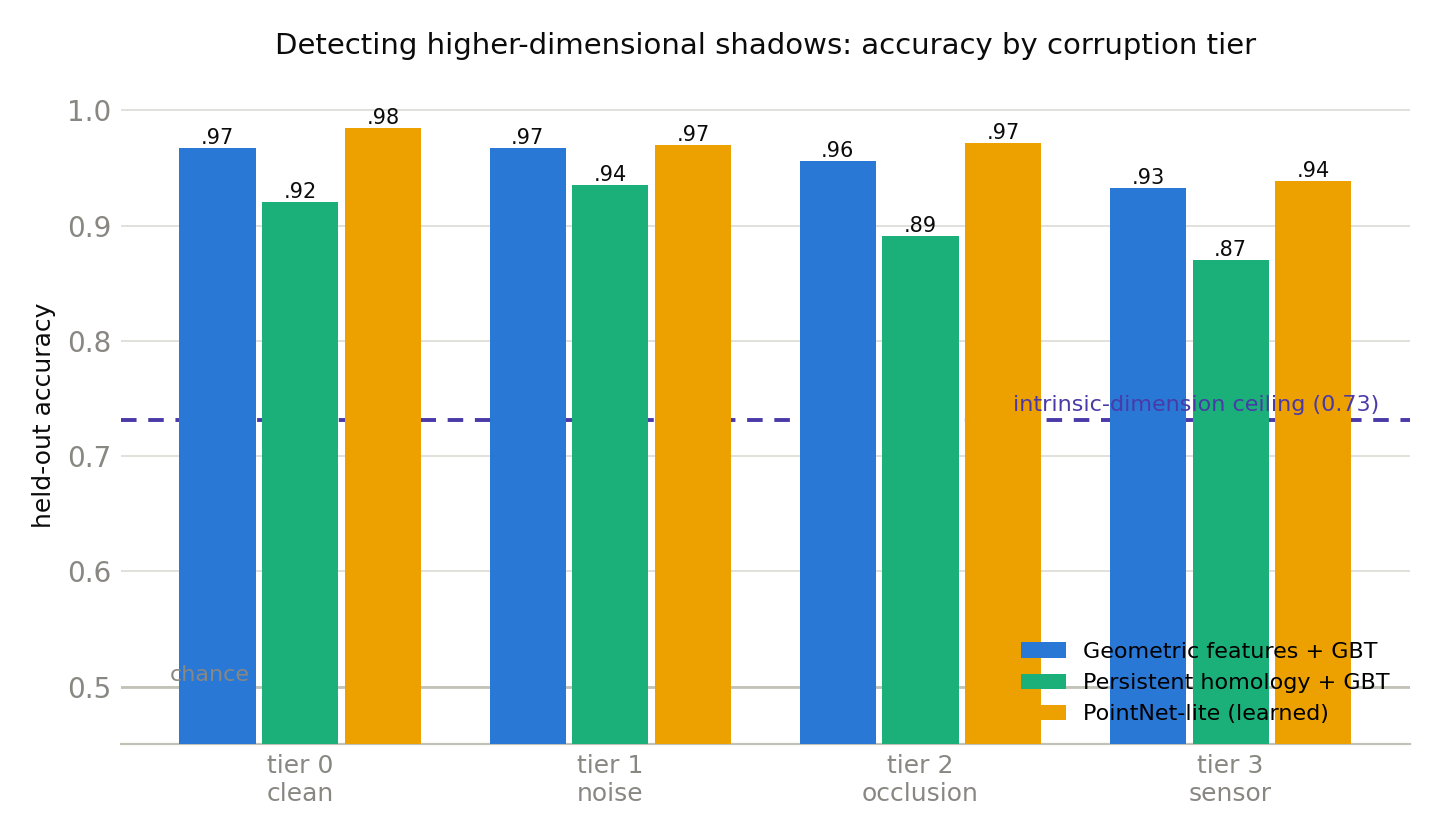}
  \caption{Accuracy by corruption tier. Both feature-based and learned methods sit far above the intrinsic-dimension ceiling (dashed) at every tier and degrade gracefully under noise, occlusion, and sensor-like dropout.}
  \label{fig:results}
\end{figure}

Table~\ref{tab:static} and Figure~\ref{fig:results} give the headline comparison. Mean estimated intrinsic dimension is 2.5 (native) vs.\ 2.9 (projected) under TwoNN, a real but small gap, confounded exactly as predicted: the solid ball (native, ID $\approx 3$) and projected 2-manifolds (ID $\approx 2$) sit on the wrong sides of any threshold. \textbf{Dimensionality is the wrong observable; the projection map, not the manifold dimension, is what leaves evidence.}

Learned and feature-based methods degrade gracefully with corruption (PointNet: 0.984 clean $\rightarrow$ 0.939 sensor tier). The per-shape error profile is theoretically coherent: the residual confusions are $S^4$-shadows vs.\ balls (0.54) and the ball class generally (0.82), pairs that differ \emph{only} in radial density profile, since the shadow of a hypersphere \emph{is} a ball with mass pushed toward the rim.

\subsection{Generalization to unseen families}

\begin{table}
  \caption{Leave-one-family-out: entire object families are excluded from training and used as the test set. Detection transfers well above chance to never-seen generators; higher ambient dimension is consistently easier.}
  \label{tab:holdout}
  \centering
  \begin{tabular}{lccc}
    \toprule
    Held-out family & Overall & 4D member & 5D member \\
    \midrule
    Hypertori ($\mathbb{R}^4 + \mathbb{R}^5$) & 0.910 & 0.82 & 1.00 \\
    Hypercubes (tesseract + 5-cube) & 0.795 & 0.64 & 0.95 \\
    \bottomrule
  \end{tabular}
\end{table}

Table~\ref{tab:holdout} shows two consistent patterns: (i) detection transfers well above chance to never-seen generators, so the model learns projection signatures rather than a shape catalog; (ii) 5D objects are uniformly easier than 4D, each collapsed dimension compounds the density-fold signature. The tesseract is the hardest object in the benchmark: its flat cells project to polyhedron-like shadows that mimic native geometry.

\subsection{Temporal track}

The rigidity witness (Section~\ref{sec:witness}) dominates: AUROC 0.982 with zero parameters, against 96.2\% for the best learned static method. \textbf{Motion reveals dimensionality that still frames hide}, consistent with, and now quantifying, the human results of \citet{he2023rigidity}.

\section{Limitations and ethics}

\paragraph{Simulation scope.} Results certify detectability under our generative assumptions (uniform sampling, two projection models, our noise models). Real sensors differ; Table~\ref{tab:holdout} measures generalization across object families but not across projection physics.

\paragraph{The witness needs correspondence.} The Kabsch witness assumes tracked points across frames. Untracked variants (e.g., distributional rigidity via optimal transport) are open.

\paragraph{Coverage.} Eleven higher-dimensional families cannot exhaust ``higher-dimensional objects''; the random-manifold family mitigates but does not eliminate this.

\paragraph{Interpretation discipline.} The benchmark will inevitably invite speculative application to anomalous real-world data. We state plainly: a positive detection on real data would establish only that the data is inconsistent with the rigid-3D model class considered, never the existence of extra dimensions. Every mundane explanation (tracking error, non-rigid objects, atmospherics, optics) dominates a priori and must be eliminated independently.

\section{Conclusion}

HyperShadow turns a century-old thought experiment, Plato's cave, Abbott's \emph{Flatland}, one dimension up, into a measurable machine-learning problem with controlled ground truth. Shadows of higher-dimensional objects are detectably different from native 3D geometry; the difference is invisible to intrinsic-dimension estimation, learnable from single frames, and certifiable from motion by a closed-form witness. We release the dataset, models, and code, and propose the systematic study of \emph{dimensional witnesses}, observable statistics bounded under low-dimensional explanations, as the path by which this line of work could eventually meet physical data.

\paragraph{Acknowledgements.} AI assistance was used in developing the synthetic data generation pipeline and baseline code, and in drafting and editing the manuscript. All experiments were designed and executed by the author, and all results, claims, and references were verified by the author.

\bibliographystyle{plainnat}

\appendix

\section{Reproducibility}

All data are generated from seeded NumPy; the following commands reproduce every number in this paper. Thirty-four unit tests verify geometric correctness (Haar orthogonality, measure-uniform sampling, projection identities).

\begin{verbatim}
python -m hypershadow.generate --out data --per-class 600 --seed 0
python -m hypershadow.generate --out data --temporal --per-class 100 --seed 1
python -m baselines.features      --data data/static.npz
python -m baselines.id_estimators --data data/static.npz
python -m baselines.tda           --data data/static.npz --max-samples 1200
python -m baselines.pointnet      --data data/static.npz --epochs 60
python -m baselines.pointnet      --data data/static.npz --epochs 40 \
    --holdout-shapes hypertorus_4d hypertorus_5d
python -m baselines.rigidity      --data data/temporal.npz
\end{verbatim}

\section{Shape gallery}

\begin{figure}[h]
  \centering
  \includegraphics[width=\linewidth]{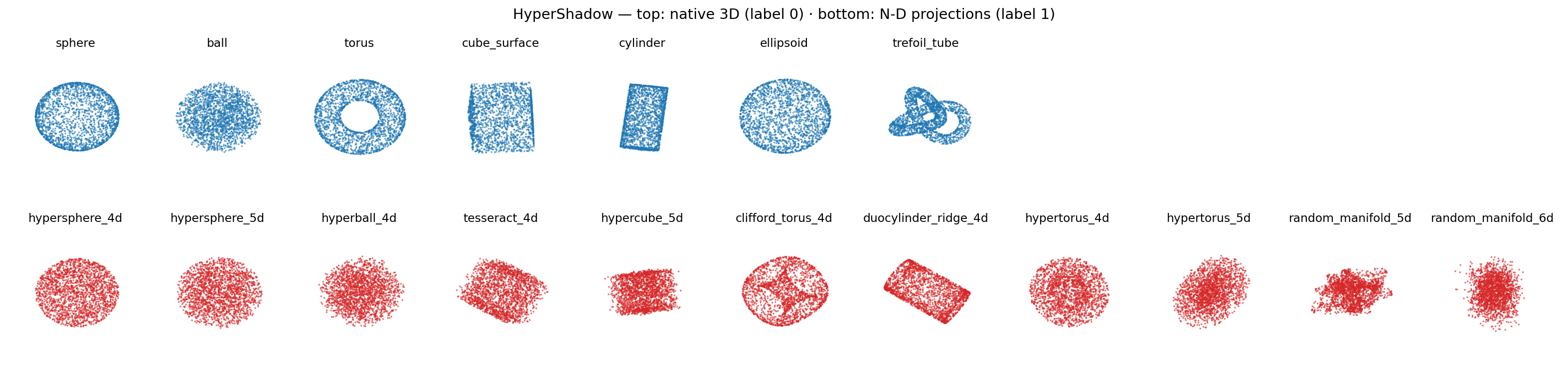}
  \caption{One sample per family (clean tier). Top: native 3D (label 0). Bottom: orthographic shadows of higher-dimensional objects (label 1). Visible signatures include the filled ball of the $S^3$ shadow, the astroid-shaped density fold of the Clifford torus, and the nested-cell structure of the tesseract.}
  \label{fig:gallery}
\end{figure}

\end{document}